\documentclass[sigconf]{acmart}
\usepackage{multirow}
\usepackage{graphicx}

\begin{document}

\title{Advancing Time Series Classification with Multimodal \\ Language Modeling}

\author{Mingyue Cheng}
\affiliation{%
  \institution{University
 of Science and Technology of China}
  \city{Hefei, Anhui Province}
  \country{China}}
\email{mycheng@ustc.edu.cn}

\author{Yiheng Chen}
\affiliation{%
  \institution{University
 of Science and Technology of China}
  \city{Hefei, Anhui Province}
  \country{China}}
\email{cyh_20@mail.ustc.edu.cn}

\author{Qi Liu*}
\affiliation{%
  \institution{University
 of Science and Technology of China}
  \city{Hefei, Anhui Province}
  \country{China}}
\email{qiliuql@ustc.edu.cn}

\author{Zhiding Liu}
\affiliation{%
  \institution{University
 of Science and Technology of China}
  \city{Hefei, Anhui Province}
  \country{China}}
\email{zhiding@mail.ustc.edu.cn}

\author{Yucong Luo}
\affiliation{%
  \institution{University
 of Science and Technology of China}
  \city{Hefei, Anhui Province}
  \country{China}}
\email{prime666@mail.ustc.edu.cn}




\renewcommand{\shortauthors}{}

\begin{abstract}

For the advancements of time series classification, scrutinizing previous studies, we can summarize that most existing methods adopt a common learning-to-classify paradigm -  a time series classifier model tries to learn the relation between sequence inputs and target label encoded by one-hot distribution. Although effective, we reveal that this paradigm conceals two inherent limitations: (1)  encoding target categories with one-hot distribution fails to reflect the comparability and similarity between labels, and (2) it is very difficult to learn transferable model across domains, which greatly hinder the development of universal serving paradigm.  In this work, we propose InstructTime, a novel attempt to reshape time series classification as a learning-to-generate paradigm. Relying on the powerful generative capacity of the pre-trained language model, the core idea is to formulate the classification of time series as a multimodal understanding task, in which both task-specific instructions and raw time series are treated as multimodal inputs while the label information is represented by texts.  To accomplish this goal, three distinct designs are developed in the InstructTime. Firstly, a time series discretization module is designed to convert continuous time series into a sequence of hard tokens to solve the inconsistency issue across modal inputs. To solve the modality representation gap issue, for one thing, we introduce an alignment projected layer before feeding the transformed token of time series into language models.  For another, before fine-tuning the language model for the target domain task, we highlight the necessity of auto-regressive pre-training across domains, which can facilitate the transferability of the language model and boost the generalization performance. Extensive experiments are conducted over benchmark datasets, whose results uncover the superior performance of InstructTime and the potential for a universal foundation model in time series classification. To facilitate further research and development in this field, we make our codes publicly available\footnote{https://github.com/Mingyue-Cheng/InstructTime}.
\end{abstract}




\maketitle

\begin{figure*}[ht]
	\centering
	\includegraphics[width=1.0\textwidth]{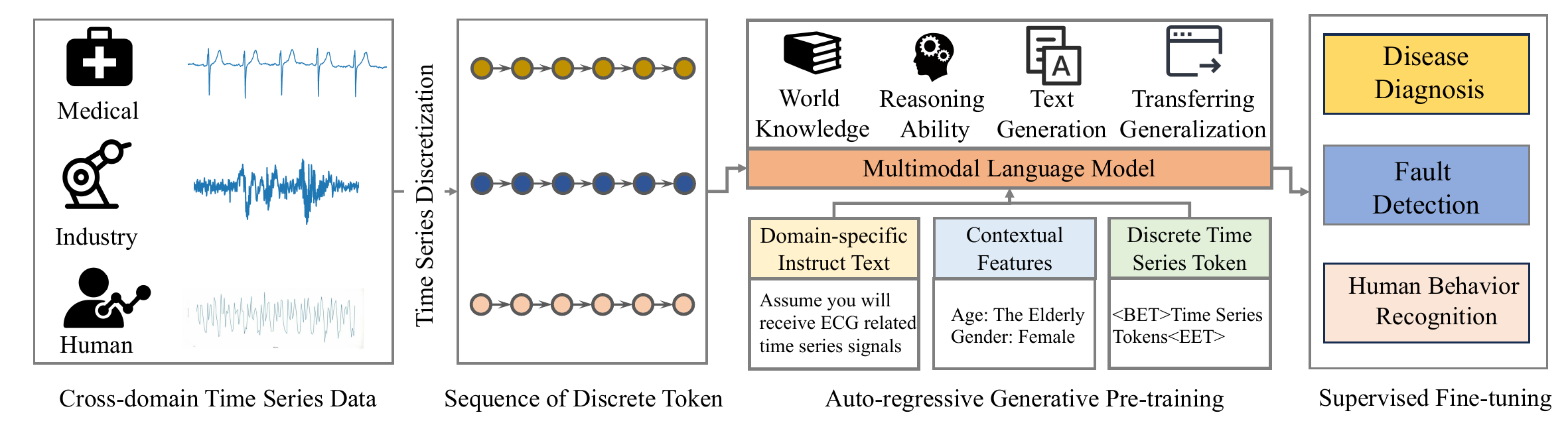}
	\vspace{-0.3in}
	\caption{Illustrating the pipeline of newly proposed InstructTime.}
	\vspace{-0.2in}
	\label{fig:example}
\end{figure*}
\section{Introduction}
Time series classification (TSC) stands as a pivotal task within the realm of data science research and has seen considerable growth over recent decades~\cite{zheng2014time,cheng2023formertime, cheng2024convtimenet}. This growth is driven by the increasing demand across a myriad of applications, from diagnosing diseases in the healthcare sector to detecting faults in the industrial internet. At its core, TSC involves the categorization of time series data into predefined classes, based on the specific context or patterns inherent in the data sequences. 

Leveraging the remarkable representation ability of deep learning, numerous feature-based TSC models~\cite{middlehurst2023bake} have emerged as significant contributors to the field. Notably, these methods excel in scalability, adeptly handling large-scale time series data across various scenarios without substantial constraints related to the increasing number of channels or instances. A thorough examination of recent advancements reveals that a majority of these models adhere to a unified learning-to-classify framework. Specifically, the primary objective of a time series classifier within this paradigm is to learn the mapping between continuous sequence inputs and their corresponding target labels, which are typically represented as one-hot vectors. This approach underscores the models' capacity to distill and categorize complex temporal patterns into discernible classes, facilitating their application across diverse domains.

While effective, we identify several inherent limitations within this learning paradigm that merit consideration. First, the paradigm presents substantial challenges for cross-domain knowledge transfer~\cite{li2023frozen,kamarthi2023large}. The variability in the number of channels across datasets from different domains can significantly impede the transferability of tasks. For example, datasets from various domains may differ in the number of channel features, complicating the direct application of a model trained in one domain to another. Second, the widespread use of one-hot encoding for target label often neglects the intrinsic similarities and comparability among categories~\cite{hinton2015distilling,cheng2021learning}. For instance, the well-known field of Human Activity Recognition (HAR), where activities such as walking and jogging bear more resemblance to each other than either does to lying down. The orthogonal nature of one-hot encoded labels does not account for these subtle inter-class relationships, which could limit the model's ability to utilize these similarities for enhanced generalization and accuracy. Hence, the exploration of alternative representation schemes that capture the nuanced similarities between categories is crucial for improving the efficacy of deep learning models. Third, the current models struggle to incorporate knowledge from rich side information effectively~\cite{rendle2010factorization,cao2023tempo}. In healthcare contexts, for example, auxiliary information such as a patient's age and gender is crucial for accurately identifying her symptoms. 

To address the aforementioned challenges, we are motivated to explore a promising yet more complex learning paradigm for the TSC task. Drawing inspiration from the recent breakthroughs in pre-trained language model (PLM)~\cite{brown2020language,radford2018improving}, our objective is to reshape time series classification as a label generation task, given inputs that combine time series data and language descriptions. However, achieving this goal presents significant hurdles due to the substantial modality differences between time series and language models. Firstly, time series data, unlike textual data, lacks a unified standard across different domains~\cite{cheng2023timemae,nie2022time,liu2024generative}. Even assuming that time series datasets from various domains could be standardized, addressing the modality alignment challenge is crucial. This is because the profound disparities between the representations of time series data and language tokens. Furthermore, the potential of employing the PLM as the backbone network to learn the mapping between text inputs and target labels is still an under-explored territory. The feasibility and effectiveness of leveraging language models for this purpose, considering their inherent design to process and generate textual content, raises intriguing questions about their adaptability to handle the nuanced dynamics of time series data. 

To overcome the identified challenges, we conceptualize the classification of time series as a multimodal understanding task and introduce a novel approach, named InstructTime. The core principle involves utilizing domain-aware instructions and time series-derived features as prompts to directly generate the desired label text, as illustrated in Figure~\ref{fig:example}. In InstructTime, we address the challenge of processing time series data with pre-trained language models by employing a discretization strategy that relies on vector quantization networks. To seamlessly bridge the gap between modalities, we further integrate an aligned projector layer implemented through a multilayer perceptron (MLP). InstructTime adopts a dual-phase tuning process: initially, it involves comprehensive fine-tuning of the PLM using multimodal inputs in an auto-regressive pre-training fashion across various domains. Subsequently, we fine-tune the tuned PLM specifically to enhance its adaptability to the target domain. These efforts ensure that InstructTime not only effectively mitigates the modality representation gap but also integrates knowledge from diverse domains efficiently. Our extensive experimentation across benchmark datasets validates the efficacy of InstructTime in numerous scenarios. 
\begin{figure*}[ht]
	\centering
	\includegraphics[width=1.0\textwidth]{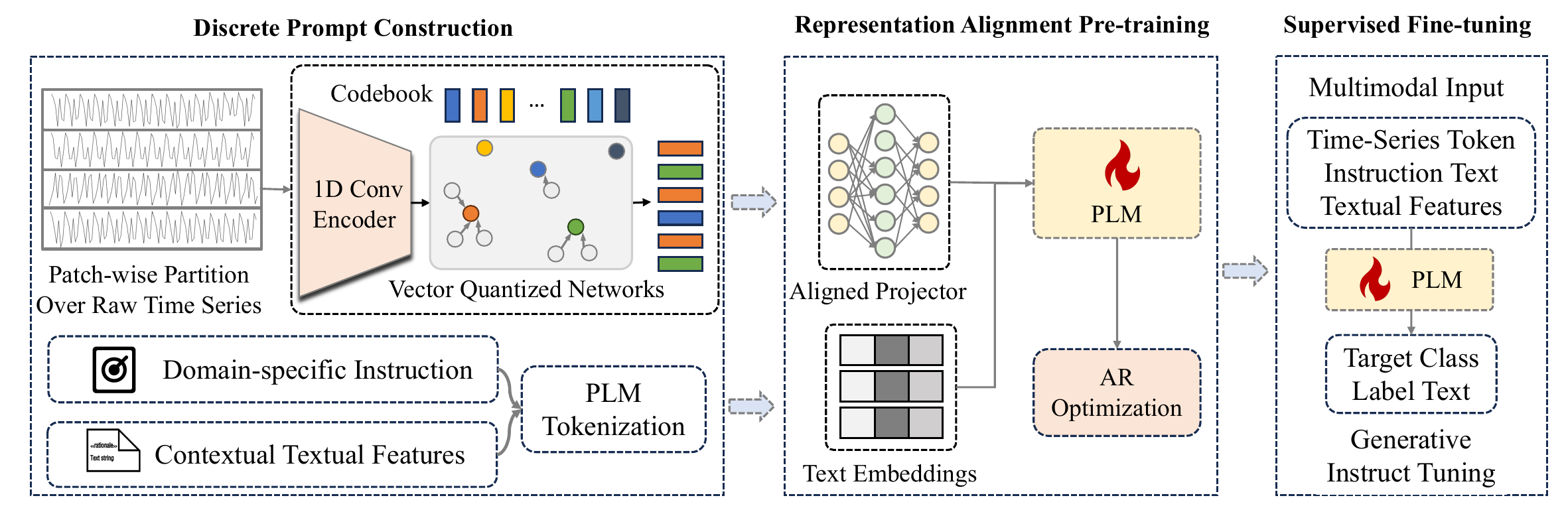}
	\vspace{-0.2in}
	\caption{Illustration of the network architecture of the InstructTime.}
	\vspace{-0.1in}
	\label{fig:frameowkr}
\end{figure*}
\vspace{-0.12in}
\section{Preliminaries}
In this section, we introduce the problem definition, basic concepts, and notations used throughout this paper. Note that due to the limitation space, we describe the close related work in Appendix~\ref{sec:related_work}.
\subsection{Problem Definition}

\paragraph{Cross Domain Time Series Representation Transferring} 
In the context of cross-domain time series representation learning, we consider a scenario involving $N$ distinct domains, each containing time series data. Let us denote these domains as $\mathcal{D} = {D_1, D_2, \ldots, D_N}$. Each domain $D_i$ comprises $M_i$ time series samples, such that $D_i = {\mathbf{x}_{1}^i, \mathbf{x}_{2}^i, \ldots, \mathbf{x}_{M_i}^i}$, where each time series $\mathbf{X}_{ij} \in \mathbb{R}^{L \times H}$ is characterized by a length of $L$ time steps and $H$ channels. The challenge in cross-domain time series representation learning lies in effectively leveraging the temporal data across these diverse domains. The goal is to harness transfer learning techniques and other related methodologies to facilitate the learning of robust time series representations. These representations should ideally encapsulate the underlying temporal dynamics and variations across domains, thereby enabling improved performance in time series classification tasks. Formally, the objective is to learn a representation function $f: \mathbb{R}^{L \times H} \rightarrow \mathbb{R}^d$, which maps a time series $\mathbf{X}_{ij}$ to a $d$-dimensional feature vector $\mathbf{x}_{ij}$, such that $\mathbf{X}_{ij} = f(\mathbf{X}_{ij})$. This function should be designed to preserve universal enough across all target domains.
\paragraph{Domain-specific Time Series Classification}
Building upon the foundation of cross-domain time series representation learning, we delve into the problem of time series classification. Within this framework, the task at hand involves categorizing each time series, represented by $\mathbf{X}_{ij} \in \mathbb{R}^d$ for the $j$-th sample in the $i$-th domain, into one of $C$ predefined categories. These categories are denoted by $\mathcal{C} = \{c_1, c_2, \ldots, c_C\}$. The classification function, denoted as $g: \mathbb{R}^d \rightarrow \mathcal{C}$, aims to assign a category label $y_{ij} \in \mathcal{C}$ to each time series representation $\mathbf{X}_{ij}$. The objective is to optimize $g$ such that it achieves high accuracy for the target domain.

\vspace{-0.12in}
\section{Methodology}
In this section, we will introduce the InstructTime, in detail. The core idea of is to treat the classification of time series as multi-modal understanding problem, where local region of  time series and language word are seen as two distinct modalities. By this way, different time series task can be easily unified together via formulating all classification task as the next token prediction task. 


\subsection{Framework Overview}
Figure~\ref{fig:frameowkr} illustrates the model architecture of InstructTime, which consists of three key stages.  The process begins with a patch-wise partitioning of time series data, encoding it through Vector Quantized Networks to produce a discrete representation. Domain-specific instructions are integrated to enrich the representation with contextual information. These enriched sequences are tokenized and aligned with text embeddings, preparing the data for the pre-trained language model (PLM). The PLM undergoes auto-regressive optimization across multiple domains. Finally, supervised fine-tuning with multimodal inputs—including time series tokens and textual instructions—tailors the PLM for specific analytical tasks, employing generative instruct tuning to enhance the model's predictive capabilities for domain-specific time series data. In the following, we introduce some key techniques of this architecture

\subsection{Discrete Prompt Construction}
\subsubsection{Time Series Discretization with Vector Quantized Networks} 
Time series data fundamentally differs from linguistic text, as it consists of sequences of continuous numerical values rather than discrete verbal or written elements. This distinction introduces unique challenges in processing and analyzing time series data. On one hand, the semantic information inherent in each time step of a time series is relatively sparse~\cite{cheng2023formertime}. Unlike words in a sentence that carry rich and distinct meanings, numerical values in time series might not individually convey significant information until they are considered as part of larger patterns or trends over time. On the other hand, the length of time series data can be considerably extensive~\cite{zhou2021informer}, often encompassing thousands to millions of data points to represent comprehensive temporal dynamics. This characteristic poses substantial challenges when attempting to directly input such data into models originally designed for processing language.  

Due to directly applying language models to time series data without accommodating these intrinsic differences can lead to significant challenges, underscoring the need for specialized approaches or adaptations that can effectively handle the unique properties of time series data. In this work, we decide to adopt a simple and direct idea-time series discretization. It should be noted that some previous related works have been proposed, like well-known Symbolic Aggregate approXimation (SAX)~\cite{lin2007experiencing}. However, we argue that these conventional approaches exhibit significant limitations that can impact their efficacy and applicability in complex real-world scenarios.  A primary concern is the substantial information loss during the compression process, where the dimensionality reduction inherent in these methods often leads to the omission of critical nuances and patterns within the time series data. This compression loss can significantly affect the subsequent analysis, leading to less accurate or insightful results. 

Based on above analysis, we decide to design a novel technique to perform time series discretization. Specifically,  the techniques of vector quantized (VQ) networks~\cite{van2017neural,gray1984vector} to perform time series discretization. The core idea is to assign each local region of sub-series with its own identity code based on reconstruction optimization over auto-encoder architecture.  The specif code is selected from a pre-defined codebook embedding space.  In this work, we employ the well-known TCN model~\cite{oord2016wavenet} as backbone architecture of encoder and decoder. Generally, we can measure the connection between local sub-series and the candidate codes with classical similarity computation function, e., squared distance. The objective is to find the appropriate discrete sequence that minimizes the summed distances. 

Formally, for $i$-th domain, we assign a trainable embedding matrix, denoted by $E_i =\{e_k^i\}_{k=1}^K = \{e_1^i, e_2^i., \ldots,  e_K^i\}$ with $K$ distinct embedding vectors. The indices of these continuous vectors can be regarded as the corresponding  discrete codes.  Before feeding the raw time series into encoder of VQ networks, the patch embedding operation is firstly performed.  Formally, given a time series instance $X_{ij} = \{z_1, z_2, \ldots, z_T \}$ in the $i$-th domain and $j$-th example,  it will be segmented into non-overlap contiguous patches $(s_1, s_2, ..., s_P)$.  Here, each patch $s_p$ denotes a sub-sequence of $z_{1:T}$, consisting of $|s_p|$ time points of raw time series instance.  By leveraging a weight-sharing 1-D convolution operation, the whole segment $s_p$ is transformed to a single representative vector. 

In the forward pass of the VQ networks, the later discretizes a sequence of continuous feature vector $(z_1, z_2,\ldots, z_P)$ by mapping each $z_p$ to its nearest neighbour in the codebook, i.e., each $z_m$ is replaced with $e_k$, where $k = \arg \text{min} \parallel z_p - e_k\parallel^2$. The layer output is the resulting quantized sequence $(\hat{z}_{s_1}, \hat{z}_{s_2}, \ldots, \hat{z}_{s_P})$. For example, using this notation, we might have $s_1$ = $z_{1:10}$ with $\hat{z}_{s1} = e_{12}$, meaning that vectors of segment $s_1$ is assigned to the code $12$. 

To train the entire VQ networks (including encoder, decoder, and codebook embedding),  the final reconstruction loss is used. Formally, it can be described as:
\vspace{-0.1in}
\begin{equation}
	\label{loss}
	\mathcal{L} = \text{log} p(x|z_q(x)) + ||\text{sg}[z_e(x)] - e||^2_2  + \beta ||z_e(x) - \text{sg} [e]||^2_2,	
\end{equation}\noindent  where $\text{sg}$ denotes stop gradient operator that is defined as identity at forward computation time and has zero partial derivatives, thus effectively constraining its operand to be a non-updated constant.  This equation involve three components in training the different part of the VQ networks. The first term is reconstruction loss, whose main role is to optimize the encoder and decoder. In addition to the reconstruction term,  this loss involve another two terms.  For the forward pass, the $\arg \text{min}$ operator is not differentiable. Hence, gradient are approximated using the straightthrough estimator (STE)~\cite{bengio2013estimating} in the backward pass. Correspondingly, the codebook is trained using exponential moving average of the continuous features as illustrated in the second terms of Equation~\ref{loss}.  Finally, since the volume of the embedding space is dimensionless, it can grow arbitrarily if the embeddings $e_k$ do not train as fast as the encoder parameters. To make sure the encoder commits to an embedding and its output does not grow, a third term, named commitment loss, is also further added. 


\subsubsection{Prompt Template Design and Highlighted Guidelines}
\begin{figure}
	\centering
	\includegraphics[width=0.48\textwidth]{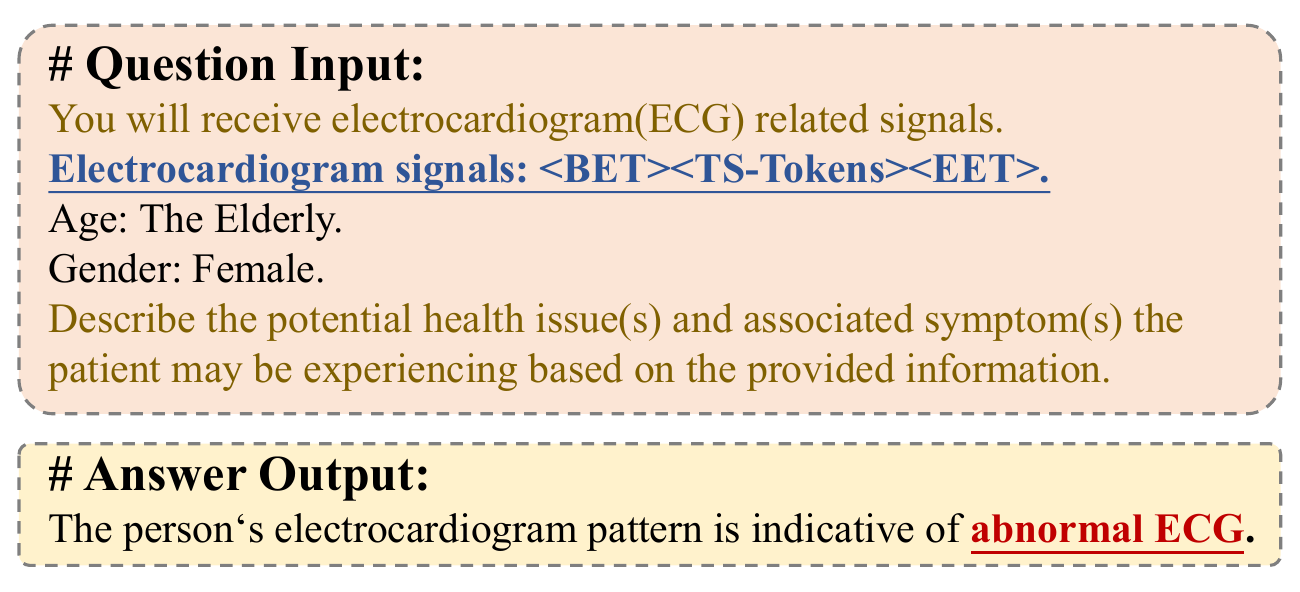}
	\vspace{-0.3in}
	\caption{Illustration of the newly proposed ConvTimeNet.}
	\vspace{-0.3in}
	\label{fig:prompt}
\end{figure}

For each domain, we utilize fixed prompt templates for target label generation. Ultimately, our fixed prompt template is structured as shown in Figure~\ref{fig:prompt}. This template totally contain three sides of information: domain-specific task description, contextual features, and discrete time series token.  Particularly, ``TS-Token'' denotes the sequence of discrete time series token projected by VQ networks while ``<BET>'' and ``EET'' denote the placeholder. It should be noted that prompt template play a vital role in language generation task. In this work, we highlight the following three guidelines in designing prompt template: (1) Inclusion of candidate label information: we incorporate candidate label information within the prompt to provide explicit cues to the PLM. This approach primes the LM to consider these labels as part of the solution space, thereby enhancing its ability to associate given inputs with potential outcomes. (2) Consistency in sequence length across domains: Given the multi-domain nature of our task, it is crucial to maintain uniform sequence lengths for time series across different domains. This consistency ensures that the LM is not biased by varying lengths and can learn domain-invariant features, facilitating better generalization and robust classification. (3) Completion of predictive answer output: Instead of merely predicting discrete answers, we instruct the model to generate a complete sentence that encapsulates the prediction. Noted that we provide sufficient experimental evidence in our experiments as illustrated in Appendix~\ref{sec:more_results}. 



\subsection{Hybrid Encoding}
Considering the significant representation gap between the modalities of time series tokens and textual words, we have implemented a hybrid encoding strategy to convert the input prompt into latent vectors, namely embeddings. As depicted in \autoref{fig:frameowkr}, for all textual content, we leverage the pre-trained language model's  native tokenization and embedding mechanisms to transform the content into tokens and their corresponding embeddings.

During prompt construction, it is crucial to ensure that the embeddings of both modalities contribute positively to model performance. Therefore, we have additionally developed an aligned projector module to mitigate any inconsistency issues. Specifically, the aligned projector is fine-tuned to convert the vector quantized representations of time series data into a format that is coherent with the embeddings used by the pre-trained language model (PLM). This alignment is facilitated through a training strategy that coalesces the time series and textual embeddings within a unified latent space. For practical implementation, a straightforward multi-layer perceptron (MLP) can be employed for this purpose.

Once the inputted prompt has been converted into a sequence of embeddings, the language model is poised to generate predictions. The PLM employs its pre-trained neural network architecture to process the sequence, utilizing the contextual information encoded within the embeddings to inform its predictions. This processing involves the traversal of the sequence through multiple layers of the PLM, each adding a level of abstraction and understanding, ultimately leading to the final output layer. Here, the PLM synthesizes the information, applying its learned patterns and dependencies to produce a prediction. This prediction not only reflects the model's interpretation of the time series data but also integrates the semantic richness of the accompanying textual information, resulting in a multimodal inference that is both precise in its classification of time series.
\vspace{-0.1in}
\subsection{Training Procedure}
\label{training_phase}
We now consider how to train the model parameters. In our work, we adopt full fine-tuning strategies during training of the pre-trained model due to the small scale of the used PLM (i.e., GPT2~\cite{radford2019language}). Functionality speaking, there exist two key training phases: (1) auto-regressive pre-training across domains; and (2) supervised fine-tuning within a domain.  

During the first training phase, the auto-regressive generative pre-training~\cite{radford2018improving} is designed to further align the representation between discrete time series tokens and language words. This is achieved through cross-domain generative pre-training, which serves a dual purpose. Firstly, it enhances the alignment of representations, ensuring that the time series tokens are embedded in a manner that is cognitively coherent with the language model's understanding of natural language words. Secondly, through cross-domain self-supervised pre-training, the model's domain generalizability is significantly improved. This phase leverages the diversity of data across domains to teach the model to recognize and adapt to underlying patterns and structures that are common to various time series datasets, thus enriching the model's ability to handle domain-specific nuances during the subsequent fine-tuning phase.

In the second phase, domain-specific supervised generative fine-tuning~\cite{ouyang2022training} is employed to tailor the pre-trained base model to the target domain effectively. Noted that we do not calculate the regression loss for this specific input prompt due to the consideration of domain alignment. By focusing on a particular domain, the model is fine-tuned with labeled data that is representative of the target tasks' specific characteristics and requirements. This phase is critical as it allows the model to apply the generalized knowledge acquired during the pre-training phase to the specialized context of the target domain, enhancing its predictive performance and accuracy. The fine-tuning adjusts the model parameters to optimize for domain-specific features, leading to a more refined model that is better equipped to handle the intricacies and particularities of the target domain's time series data.

\section{Experiments}
In this section, we first assess InstructTime on different datasets, comparing it to other baseline approaches. Then, we conduct experiments to explore how to stabilize the performance of InstructTime. At last, we show a \textit{case study} to visualize the outcomes at each stage of InstructTime's training process.

\subsection{Experiment Settings}

In our experimental setup, we undertake comprehensive evaluations across five widely recognized time series classification benchmark datasets: EEG (Electroencephalogram), ECG (Electrocardiogram), HAR (Human Activity Recognition), FD (Fault Detection), and RWC (Real-World Complex). For an in-depth description of these datasets, readers are directed to the supplementary material provided in Appendix~\ref{sec:data}. To ascertain the efficacy of our proposed InstructTime framework, we benchmark it against a diverse array of competitive baselines encompassing various modeling paradigms: (1) Self-attention based models including TST, Patch-TST, and FormerTime~\cite{cheng2023formertime}; (2) Convolutional based approaches such as MCDCNN~\cite{zheng2014time}, TCN~\cite{oord2016wavenet}, and MiniROCKET~\cite{dempster2021minirocket}; (3) Self-supervised learning models like TimeMAE~\cite{cheng2023timemae} and TS-TCC~\cite{eldele2021time}; and (4) a novel adaptation of GPT2, where its final layer is modified for classification purposes, herein referred to as GPT-As-Classifier. Furthermore, we evaluate our methods in two settings: InstructTime-Universal and InstructTime-Adapt. The former means no domain-specific fine-tuning (as stated in section~\ref{training_phase}) is performed while the latter indicates that both cross-domain and domain-specific fine-tuning are performed. That is, the InstructTime-Universal model indicate a truly foundation model which can be used to serve different domains while the latter is only used to serve the target domain. The performance of InstructTime and these baselines is quantitatively assessed using Accuracy and F1 Score as the primary evaluation metrics, ensuring a comprehensive understanding of each model's predictive capabilities. For readers interested in the specific model configurations and hyper-parameter settings employed in our experiments, we provide detailed information in Appendix~\ref{sec:implement_appendix}, facilitating reproducibility and further exploration of our findings.

\begin{table*}[htbp]
	\centering
	\caption{Classification results of all compared methods over five benchmark datasets.}
	\vspace{-0.15in}
		\resizebox{0.975\textwidth}{!}{%
	\begin{tabular}{c|cccccccccc}
					\toprule
		\multirow{2}[3]{*}{Compared Models} & \multicolumn{2}{c}{EEG} & \multicolumn{2}{c}{ECG} & \multicolumn{2}{c}{HAR} & \multicolumn{2}{c}{FD} & \multicolumn{2}{c}{RWC} \\
		\cmidrule{2-11}          & Accuracy & F1 Score & Accuracy & F1 Score & Accuracy & F1 Score & Accuracy & F1 Score & Accuracy & F1 Score \\
		\midrule
		Transformer & 0.7940  & 0.5178  & 0.1821  & 0.3691  & 0.9148  & 0.9149  & 0.9451  & 0.9564  & 0.7158  & 0.7152  \\
		Patch Transformer & 0.8076  & 0.5460  & 0.2465  & 0.3883  & 0.8704  & 0.8683  & 0.9390  & 0.9471  & 0.7552  & 0.7545  \\
		FormerTime & 0.8356  & 0.5828  & 0.3712  & 0.5233  & 0.9199  & 0.9198  & 0.9732  & 0.9853  & \textbf{0.7803 } & \textbf{0.7796 } \\
		\midrule
		MCDCNN & 0.8102  & 0.5395  & 0.0929  & 0.1735  & 0.8873  & 0.8862  & 0.9396  & 0.9545  & 0.7762  & 0.7759  \\
		TCN   & 0.7525  & 0.3927  & 0.1014  & 0.1654  & 0.9002  & 0.8997  & 0.7962  & 0.7248  & 0.7113  & 0.7109  \\
		MiniROCKET & 0.8318  & 0.5638  & 0.2689  & 0.3900  & 0.9173  & 0.9153  & 0.9412  & 0.9569  & 0.7569  & 0.7556  \\
		\midrule
		TimeMAE & 0.8248  & 0.5865  & 0.2546  & 0.3834  & 0.9294  & 0.9284  & 0.9878  & 0.9904  & 0.7690  & 0.7664  \\
		TS-TCC & 0.7291  & 0.4347  & 0.1778  & 0.3780  & 0.8832  & 0.8815  & 0.9296  & 0.9363  & 0.6979  & 0.6931  \\
		\midrule
		GPT-As-Classifier & 0.7689  & 0.4929  & 0.2253  & 0.3557  & 0.8973  & 0.8963  & 0.9489  & 0.9598  & 0.7554  & 0.7553  \\
		\midrule
		InstructTime-Universal & 0.8067  & 0.5007  & 0.3402  & 0.4820  & 0.8990  & 0.8944  & 0.9619  & 0.9656  & 0.7307  & 0.7299  \\
		InstructTime-Adapt & \textbf{0.8452 } & \textbf{0.6240 } & \textbf{0.4121 } & \textbf{0.5547 } & \textbf{0.9298 } & \textbf{0.9307 } & \textbf{0.9901 } & \textbf{0.9917 } & 0.7599  & 0.7578  \\
		\bottomrule
	\end{tabular}%
}
	\vspace{-0.12in}
	\label{tab:main}%
\end{table*}%

\begin{table*}[htbp]
	\centering
	\caption{Results of our InstructTime model with and without auto-regressive pre-training across domains}
	\vspace{-0.15in}
	\resizebox{0.975\textwidth}{!}{%
		\begin{tabular}{cc|cccccccccc}
			\toprule
			\multicolumn{1}{c}{\multirow{2}[2]{*}{\begin{tabular}{@{}c@{}}Models\end{tabular}}} & \multirow{2}[2]{*}{Settings} & \multicolumn{2}{c}{EEG} & \multicolumn{2}{c}{ECG} & \multicolumn{2}{c}{HAR} & \multicolumn{2}{c}{FD} & \multicolumn{2}{c}{RWC} \\
			\cmidrule{3-12}                            &       & Accuracy & F1 Score & Accuracy & F1 Score & Accuracy & F1 Score & Accuracy & F1 Score & Accuracy & F1 Score \\
			\midrule
			\multirow{2}[2]{*}{InstructTime-Universal} & w/o Cross-Domain & 0.7854  & 0.4854  & 0.2554  & 0.3751  & 0.8341  & 0.8296  & 0.9092  & 0.9203  & 0.7270  & 0.7268  \\
			& w/ Cross-Domain & \textbf{0.8067 } & \textbf{0.5007 } & \textbf{0.3402 } & \textbf{0.4820 } & \textbf{0.8990 } & \textbf{0.8944 } & \textbf{0.9619 } & \textbf{0.9656 } & \textbf{0.7307 } & \textbf{0.7299 } \\
			\midrule
			\multirow{2}[2]{*}{InstructTime-Adapt} & w/o Cross-Domain & 0.8283  & 0.5751  & 0.3740  & 0.5209  & 0.9019  & 0.9028  & 0.9813  & 0.9830  & 0.7482  & 0.7478  \\
			& w/ Cross-Domain & \textbf{0.8452 } & \textbf{0.6240 } & \textbf{0.4121 } & \textbf{0.5547 } & \textbf{0.9298 } & \textbf{0.9307 } & \textbf{0.9901 } & \textbf{0.9917 } & \textbf{0.7599 } & \textbf{0.7578 } \\
			\bottomrule
		\end{tabular}%
	}
		\vspace{-0.12in}
	\label{tab:cross_domain}%
\end{table*}%

\subsection{Experimental Results}
\subsubsection{\textbf{Classification Results Comparison}}
Table~\ref{tab:main} presents a comprehensive comparison of various model's classification performance over five datasets. From the results, it is clear that our InstructTime-Adapt model generally outperforms other models across most datasets, achieving the optimal classification performance.  This results indicates a robustness in the InstructTime-Adapt model, which seems to be well-suited for a variety of classification tasks, adapting effectively to different data distributions. Meanwhile, our InstructTime-Universal model also shows strong performance, especially in the FD dataset, highlighting the potential of foundation time series model. The remain baseline models show competitive results but do not consistently reach the performance level of the InstructTime models. For instance, the Transformer and Patch Transformer are strong in the EEG and ECG datasets but less effective in HAR and FD compared to the InstructTime approaches. Overall, the InstructTime models, particularly the InstructTime-Adapt variant, demonstrate superior performance across multiple datasets. This suggests that the methodologies implemented in these models are effective for handling the complexities and variations inherent in time-series classification tasks. 

\begin{table}[t]
	\centering
	\caption{Transfer learning results of external two domain with pre-trained InstructTime-Adapt on five datasets.}
	\vspace{-0.15in}
	\begin{tabular}{ccccc}
		\toprule
		\multirow{2}[2]{*}{Models} & \multicolumn{2}{c}{Epilepsy} & \multicolumn{2}{c}{AD} \\
		& Accuracy & F1 Score & Accuracy & F1 Score \\
		\midrule
		w/o Pre-training & 0.8896  & 0.5558  & 0.8045  & 0.7199  \\
		w/ Pre-training & \textbf{0.9596 } & \textbf{0.9355 } & \textbf{0.9237 } & \textbf{0.9234 } \\
		\bottomrule
	\end{tabular}%
	\label{tab:non-overlap}%
		\vspace{-0.22in}
\end{table}%

\begin{table*}[htbp]
	\centering
	\caption{Performance comparison of InstructTime-Adapt in terms of w/o or w/ auto-regressive pre-training stage.}
	\vspace{-0.1in}
	\begin{tabular}{c|cccccccccc}
		\toprule
		\multirow{2}[4]{*}{Model Variants} & \multicolumn{2}{c}{EEG} & \multicolumn{2}{c}{ECG} & \multicolumn{2}{c}{HAR} & \multicolumn{2}{c}{FD} & \multicolumn{2}{c}{RWC} \\
		\cmidrule{2-11}          & Accuracy & F1 Score & Accuracy & F1 Score & Accuracy & F1 Score & Accuracy & F1 Score & Accuracy & F1 Score \\
		\midrule
		w/o Pre-training & 0.7854  & 0.4854  & 0.2554  & 0.3751  & 0.8341  & 0.8296  & 0.9092  & 0.9203  & 0.7270  & 0.7268  \\
		w/ Pre-training & \textbf{0.8452 } & \textbf{0.6240 } & \textbf{0.4121 } & \textbf{0.5547 } & \textbf{0.9298 } & \textbf{0.9307 } & \textbf{0.9901 } & \textbf{0.9917 } & \textbf{0.7599 } & \textbf{0.7578 } \\
		\bottomrule
	\end{tabular}%
		\vspace{-0.12in}
	\label{tab:ar}%
\end{table*}%

\begin{table*}[htbp]
	\centering
	\caption{Effect of preserving textual Information during auto-regressive training on InstructTime-Adapt.}
	\vspace{-0.1in}
	\resizebox{0.975\textwidth}{!}{%
		\begin{tabular}{c|cccccccccc}
			\toprule
			\multirow{2}[2]{*}{Model Variants} & \multicolumn{2}{c}{EEG} & \multicolumn{2}{c}{ECG} & \multicolumn{2}{c}{HAR} & \multicolumn{2}{c}{FD} & \multicolumn{2}{c}{RWC} \\
			\cmidrule{2-11}                            & Accuracy & F1 Score & Accuracy & F1 Score & Accuracy & F1 Score & Accuracy & F1 Score & Accuracy & F1 Score \\
			\midrule
			w/o Text & 0.7748  & 0.4641  & 0.1675  & 0.1779  & 0.8697  & 0.8676  & 0.8937  & 0.8984  & 0.7492  & 0.7489  \\
			w/  Text & \textbf{0.8452 } & \textbf{0.6240 } & \textbf{0.4121 } & \textbf{0.5547 } & \textbf{0.9298 } & \textbf{0.9307 } & \textbf{0.9901 } & \textbf{0.9917 } & \textbf{0.7599 } & \textbf{0.7578 } \\
			\bottomrule
		\end{tabular}%
	}
		\vspace{-0.12in}
	\label{tab:text}%
\end{table*}%

\subsubsection{\textbf{Effectiveness of Cross Domain Transferring}}
\paragraph{\textbf{Importance of Cross Domain Pre-training}}
In this part, we aim to provide insights into the impact of cross-domain pre-training on the performance of two model, including InstructTime-Universal and InstructTime-Adapt. The experiments are conducted across five different domains. As illustrated in Table~\ref{tab:cross_domain}, for both InstructTime-Universal and InstructTime-Adapt models, we observe that  incorporating cross-domain pre-training results in a notable improvement across all domains. The results strongly indicate that cross-domain pre-training can enhance the generalization capabilities of the InstructTime models, leading to better performance across diverse datasets. This enhancement is likely due to the models gaining a more robust and generalized representation of features that are beneficial across different domains, demonstrating the effectiveness of leveraging cross-domain data during pre-training.
\paragraph{\textbf{Cross Domain Transferring with Non-overlapped Domains}}
In this part, we first select two datasets Epilepsy and AD that have never been pre-trained on InstructTime to perform SFT in the pre-taining model and compare them with the original model that also does not have a pre-training stage. These two datasets are described in detail in Appendix~\ref{sec:data}, and the specific experimental results are presented in Table~\ref{tab:non-overlap}. We can see that the accuracy and F1 scores after the SFT performed by InstrcutTime are much higher than those of the original model, which indicates that InstructTime's pre-training model can show a certain degree of generalization and is capable of fine-tuning the unknown dataset for this model.  Secondly, we show a model comparison experiment that removes the pre-training stage and proceeds directly to SFT (w/o Pre-training) and the model that performs SFT after pre-training (w/ Pre-training). The experimental results show that the effect of the model in the w/ Pre-training stage completely outperforms the model in the w/o Pre-training stage. This indicates that the text and time series token alignment we performed in the pre-training stage is effective. Therefore, we believe that the second stage of pre-training is essential.  In conclusion, the pre-training stage is indispensable in the whole training step of the model because it takes the task of aligning the modalities. Moreover, the pre-training model has some generalization ability for unknown datasets.





\subsubsection{\textbf{Strength of Auto-regressive Pre-training}}
The experimental results in Table~\ref{tab:ar} vividly demonstrate the critical role of auto-regressive pre-training in enhancing the performance of the InstructTime model. Across five datasets, the variant of InstructTime-Universal with auto-regressive pre-training (w/ Pre-training) significantly outperforms the model without it (w/o Pre-training). For example, in the EEG dataset, the model with pre-training achieves an accuracy of 0.8452 and an F1 score of 0.6240, compared to the 0.7854 accuracy and 0.4854 F1 score of the model lacking pre-training. This trend is consistent across all datasets, with particularly striking improvements observed in the FD dataset, where the model with pre-training reaches an accuracy of 0.9901 and an F1 score of 0.9917, showcasing near-perfect performance. These results underline the importance of incorporating auto-regressive pre-training in the InstructTime framework. Without this crucial stage, the model's effectiveness is significantly diminished, highlighting the integral role of pre-training in preparing the model to better capture and generalize from the underlying patterns in the data.

\begin{figure}[t]
	\centering
	\includegraphics[width=0.5\textwidth]{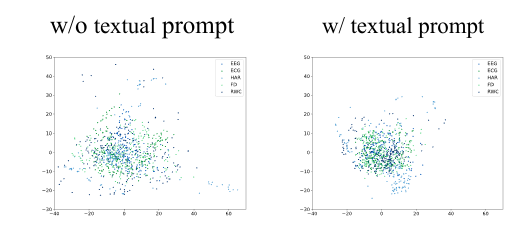}
	\vspace{-0.3in}
	\caption{T-SNE embeddings of randomly sampled instance from five datasets.}
	\vspace{-0.3in}
	\label{fig:tsne_five}
\end{figure}

\subsubsection{\textbf{In-depth Analysis of Instruction Text in Modality Alignment}}
Table~\ref{tab:text} illustrates the significant impact of incorporating textual prompts during auto-regressive training on the InstructTime-Frozen model's performance across five datasets: EEG, ECG, HAR, FD, and RWC. When comparing model variants with (``w/ Prompt'') and without (``w/o Prompt'') prompts, it is evident that the inclusion of prompts leads to substantial improvements in both accuracy and F1 scores across all datasets. For instance, in the EEG dataset, the model with prompts achieves an accuracy of 0.8452 and an F1 score of 0.6240, a notable increase from the 0.7748 accuracy and 0.4641 F1 score of the model without prompts. Similar trends are observed in other datasets, with the most significant gains seen in the FD dataset, where the model with prompts reaches near-perfect performance metrics of 0.9901 in accuracy and 0.9917 in F1 score. These results underscore the effectiveness of leveraging textual information within the auto-regressive training framework of InstructTime, enhancing the model's ability to understand and classify complex patterns, thereby boosting overall performance across diverse tasks.

We guess the reason for this result is that in high-dimensional representation spaces, if only TS tokens are used for auto-regressive pre-training, the trained embedding vectors may not be well aggregated due to the great differences in waveforms from different domains. However, if each training sample contains text, it will make all samples have a new common attribute. In this way, the samples will converge in the same direction in the auto-regressive pre-training, which leads to better aggregation and improves the generalization ability of the model. To verify this guess, we first randomly extract 200 samples from 5 datasets to form a test set of 1000 samples and feed them into two pre-training models. we extract the hidden layer vectors of both models before the classification layer and pass a mean pooling layer, and then the results are downscaled using the PCA method, resulting in a two-dimensional visualized image, as shown in Figure~\ref{fig:tsne_five}. We believe that the distribution of samples in this figure verifies our guess quite directly.

\begin{table*}[htbp]
	\centering
	\caption{Experimental results of our InstructTime-Universal tuned with different levels of training set.}
	\vspace{-0.1in}
	\resizebox{0.975\textwidth}{!}{%
		\begin{tabular}{cc|cccccccccc}
			\toprule
			\multirow{2}[2]{*}{\begin{tabular}{@{}c@{}}Model\\Variants\end{tabular}} & \multirow{2}[2]{*}{Proportion} & \multicolumn{2}{c}{EEG} & \multicolumn{2}{c}{ECG} & \multicolumn{2}{c}{HAR} & \multicolumn{2}{c}{FD} & \multicolumn{2}{c}{RWC} \\
			\cmidrule{3-12}                            &       & Accuracy & F1 Score & Accuracy & F1 Score & Accuracy & F1 Score & Accuracy & F1 Score & Accuracy & F1 Score \\
			\midrule
			\multicolumn{1}{c}{\multirow{4}[1]{*}{\begin{tabular}{@{}c@{}}w/o\\Pre-training\end{tabular}}} & 10\%  & 0.6151  & 0.3061  & 0.1467  & 0.1538  & 0.6512  & 0.6317  & 0.6979  & 0.5319  & 0.5367  & 0.3322  \\
			& 40\%  & 0.6685  & 0.3163  & 0.1318  & 0.2321  & 0.7336  & 0.7160  & 0.7309  & 0.5720  & 0.5107  & 0.3451  \\
			& 70\%  & 0.7431  & 0.4219  & 0.1882  & 0.2664  & 0.7512  & 0.7346  & 0.7639  & 0.5936  & 0.5724  & 0.4146  \\
			
			\midrule
			\multicolumn{1}{c}{\multirow{4}[2]{*}{\begin{tabular}{@{}c@{}}w/\\Pre-training\end{tabular}}} & 10\%  & 0.6974  & 0.4070  & 0.3781  & 0.5185  & 0.7774  & 0.6627  & 0.8244  & 0.6251  & 0.7044  & 0.6825  \\
			& 40\%  & 0.7368  & 0.4666  & 0.3922  & 0.5421  & 0.8357  & 0.7140  & 0.8831  & 0.6780  & 0.7253  & 0.7008  \\
			& 70\%  & 0.7607  & 0.4728  & 0.3972  & 0.5491  & 0.8571  & 0.7337  & 0.8842  & 0.9063  & 0.7080  & 0.7088  \\
			\bottomrule
		\end{tabular}%
	}
		\vspace{-0.12in}
	\label{tab:few-shot}%
\end{table*}%
\begin{table*}[htbp]
	\centering
	\caption{Experimental results of InstructTime-Universal model variants with various projection layer configurations.}
	\vspace{-0.1in}
	\resizebox{0.975\textwidth}{!}{%
		\begin{tabular}{cc|cccccccccc}
			\toprule
			\multirow{2}[2]{*}{\begin{tabular}{@{}c@{}}Alignment\\Projector\end{tabular}} & \multirow{2}[2]{*}{Hidden Size} & \multicolumn{2}{c}{EEG} & \multicolumn{2}{c}{ECG} & \multicolumn{2}{c}{HAR} & \multicolumn{2}{c}{FD} & \multicolumn{2}{c}{RWC} \\
			\cmidrule{3-12}                            &       & Accuracy & F1 Score & Accuracy & F1 Score & Accuracy & F1 Score & Accuracy & F1 Score & Accuracy & F1 Score \\
			\midrule
			Linear & 64, 768 & 0.7776  & 0.4287  & 0.2699  & 0.3410  & 0.8614  & 0.8532  & 0.9249  & 0.9310  & 0.7118  & 0.7082  \\
			\midrule
			\multirow{2}[2]{*}{MLP} & 128, 512, 768 & 0.7935  & 0.4849  & 0.3060  & 0.4451  & 0.8885  & 0.8828  & \textbf{0.9685 } & \textbf{0.9714 } & 0.7227  & 0.7121  \\
			& 64, 128, 256, 512, 768 & \textbf{0.8067 } & \textbf{0.5007 } & \textbf{0.3402 } & \textbf{0.4820 } & \textbf{0.8990 } & \textbf{0.8944 } & 0.9619  & 0.9656  & \textbf{0.7307 } & \textbf{0.7299 } \\
			\bottomrule
		\end{tabular}%
	}
	\vspace{-0.12in}
	\label{tab:projector}%
\end{table*}%

\subsubsection{\textbf{Performance Comparison of Multi-class and Multi-label Classification}}
\begin{figure}[t]
	\centering
	\includegraphics[width=0.5\textwidth]{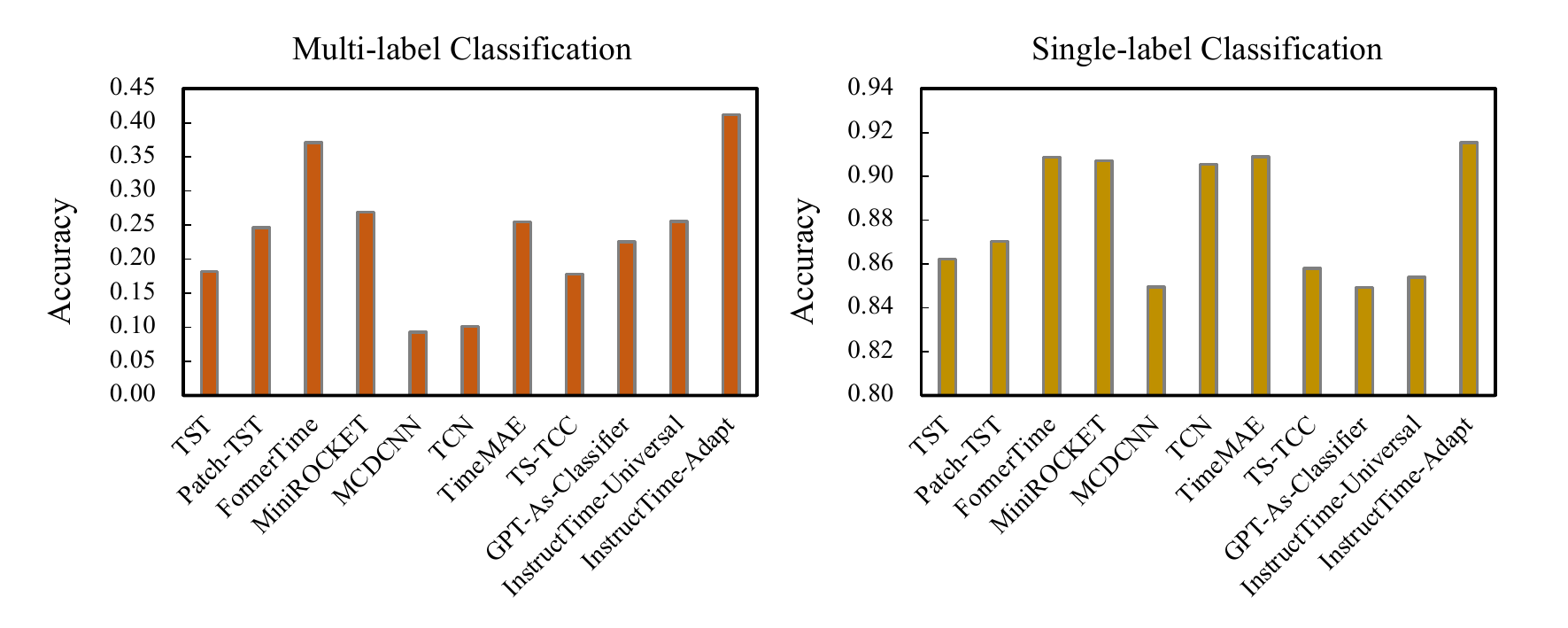}
	\vspace{-0.3in}
	\caption{Experimental results of multi-label and single-label ECG classification in terms of all compared methods.}
	\vspace{-0.2in}
	\label{fig:multi-label}
\end{figure}

To assess the performance of various algorithms on complex multi-label and simpler multi-class classification tasks, we evaluated the methods using ECG data. As illustrated in Figure~\ref{fig:multi-label}, most methods demonstrate relatively high performance on the multi-class classification task. This outcome is expected as multi-label classification is inherently more challenging. Notably, our InstructTime-Adapt method outperforms the compared methods by achieving the highest accuracy. Specifically, our method shows greater improvement over the baselines in multi-label classification, indicating its potential benefits in managing complex classification scenarios. We hypothesize that the probable reason for this improvement is that the inductive biases inherent in multi-label text can be effectively captured through our generative approach.

\subsubsection{\textbf{Data-Efficient Evaluation}}
For the data-efficient evaluations, which is also known as few-shot setting~\cite{brown2020language}, we gradually reduce the fraction of training set with $70\%, 40\%, 10\%$. This setting allow us to explore the model's ability to exhibit accuracy classification despite very limited training data, which might be particular helpful in data sparsity scenarios. The experimental results are presented in Table~\ref{tab:few-shot}, where two model variants are considered: one without pre-training and one with pre-training. As the proportion of the training set increases from $10\%$ to $70\%$, there is a general trend of improvement in all datasets. This suggest a larger amount of training data contributes to better model performance. Notably, the magnitude of improvement varies by dataset. For instance, the HAR dataset shows a more pronounced increase in performance metrics than the ECG dataset. This is reasonable since pre-trained model can provide a good Initialization.

\begin{table*}[htbp]
	\centering
	\caption{The influence of varying token number during the tokenization stage in the InstructTime-Universal method.}
	\vspace{-0.1in}
	\resizebox{0.975\textwidth}{!}{%
		\begin{tabular}{c|ccccccccccccccc}
			\toprule
			\multirow{2}[2]{*}{Num. Token} & \multicolumn{3}{c}{EEG} & \multicolumn{3}{c}{ECG} & \multicolumn{3}{c}{HAR} & \multicolumn{3}{c}{FD} & \multicolumn{3}{c}{RWC} \\
			\cmidrule{2-16}                            & MSE   & Accuracy & F1 Score & MSE   & Accuracy & F1 Score & MSE   & Accuracy & F1 Score & MSE   & Accuracy & F1 Score & MSE   & Accuracy & F1 Score \\
			\midrule
			128   & 0.0684  & 0.7847  & 0.4586  & 0.0732  & \textbf{0.2475 } & \textbf{0.3569 } & 0.0579  & 0.7766  & 0.7610  & 0.0732  & 0.8455  & 0.8369  & 0.0862  & 0.7013  & 0.6994  \\
			256   & 0.0595  & \textbf{0.7931 } & \textbf{0.5122 } & 0.0769  & 0.2442  & 0.3394  & 0.0501  & \textbf{0.8695 } & \textbf{0.8683 } & 0.0769  & 0.8462  & 0.8205  & 0.1998  & 0.7194  & 0.7183  \\
			384   & \textbf{0.0571 } & 0.7882  & 0.4911  & 0.0766  & 0.2206  & 0.3032  & 0.0459  & 0.8643  & 0.8000  & 0.0766  & 0.8613  & 0.8883  & 0.2024  & \textbf{0.7314 } & \textbf{0.7311 } \\
			512   & 0.0579  & 0.7822  & 0.4739  & 0.0711  & 0.1870  & 0.2980  & 0.0471  & 0.8439  & 0.7812  & 0.0711  & \textbf{0.9245 } & \textbf{0.9301 } & \textbf{0.1993 } & 0.7148  & 0.6003  \\
			768   & 0.0575  & 0.7945  & 0.5000  & \textbf{0.0684 } & 0.2388  & 0.3542  & \textbf{0.0433 } & 0.6861  & 0.6603  & \textbf{0.0684 } & 0.8880  & 0.8915  & 0.2117  & 0.6934  & 0.6874  \\
			\bottomrule
		\end{tabular}%
	}
		\vspace{-0.12in}
	\label{tab:token_number}%
\end{table*}%

\begin{figure*}[t]
	\centering
	\includegraphics[width=1.0\textwidth]{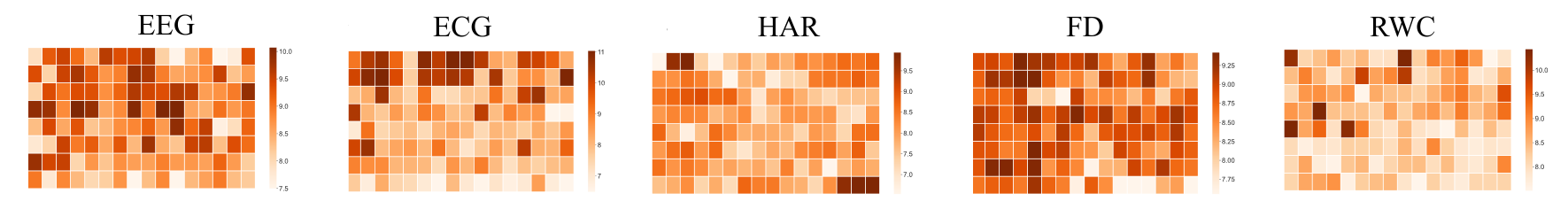}
	\vspace{-0.3in}
	\caption{Statics of used token frequency in terms of five datasets. Each heatmap corresponds to a dataset (EEG, ECG, HAR, RWC, and DEV), with color intensity indicating the frequency of token appearance.}
		\vspace{-0.2in}
	\label{fig:token}
\end{figure*}

\subsubsection{\textbf{Impacts of Alignment Projector}}
Table~\ref{tab:projector} displays the performance of InstructTime model variants with different projection layer configurations, specifically comparing a linear alignment projector against an MLP (multi-layer perceptron) with various hidden sizes. Across datasets such as EEG, ECG, HAR, FD, and RWC, it is evident that the MLP projector with a more complex hidden size configuration significantly enhances both accuracy and F1 scores. In summary, the experimental results underscore the importance of the projection layer's configuration in the InstructTime model, with a more nuanced MLP projector leading to significant performance improvements across multiple datasets. For instance, the MLP projector with hidden sizes of 64, 128, 256, 512, and 768 achieves the highest performance metrics, notably in the EEG dataset with an accuracy of 0.8067 and an F1 score of 0.5007, and in the HAR dataset with an accuracy of 0.8990 and an F1 score of 0.8944. This trend is consistent across other datasets, with the MLP projector outperforming the Linear one, highlighting the benefit of employing a more complex projection layer in capturing and processing the data's underlying patterns more effectively.

\subsubsection{\textbf{Study of Time Series Discretion}}
The experimental results depicted in \autoref{tab:token_number} illustrate the intricate relationship between the number of tokens used during the tokenization stage of the InstructTime method and its subsequent performance across multiple datasets. The performance metrics utilized for this analysis include mean squared error (MSE), Accuracy, and F1 Score, offering a holistic view of the model's effectiveness in various contexts. Across all datasets, there is a discernible trend where neither the smallest nor the largest token counts consistently yield the best results. Instead, intermediate token sizes often strike an optimal balance, enhancing the model's ability to accurately represent and process the data without succumbing to the pitfalls of overfitting or computational inefficiency. This suggests that there exists a "sweet spot" in token numbers that effectively captures the necessary information granularity while maintaining model tractability.

The experiment results in Figure~\ref{fig:token} presented in the heatmap visuals show the frequency of token usage across five different domains. From the results, we observe that the distribution of token usage varies across datasets. Some datasets like HAR, and RWC show more uniform usage of tokens across the board, indicated by a more homogenous color distribution. This suggests that the data in these domains might have a more balanced variety of features or states captured by the tokens. In contrast, datasets like EEG, ECG and FD exhibit more variance in token usage, with certain tokens appearing more frequently than others, as indicated by patches of darker colors. This could imply that certain states or features are more prevalent or that there's a higher degree of redundancy in the data within these domains. In summary, the token frequency heatmaps suggest that different domains have varying levels of feature diversity and redundancy.

\section{Conclusion and Limitations}
In this work, we propose InstructTime, attempting to reshaping the classification of time series as a learning-to-generating task, instead of a learning-to-classify as previous works do. By formulating the time series classification as a multimodal language understanding problem, InstructTime can generate the corresponding labels of each instance input. Relying on the designed time series discretion operation, InstructTime can makes it possible to transfer model parameters among different domains.  Moreover, InstructTime also introduce several key technique settings to solve the representation inconsistency and gap issues.  Experimental results show that InstructTime bring great improvements to the time series classification task, indicating its potential impact for this area. 

Meanwhile, we also realize that there exist some limitations: (1) one of the strength of the InstructTime is the powerful capacity of modeling rich contextual information~\cite{cao2023tempo}, which has not be verified in this work. An potential direction for future research would be to extend the evaluation of our methods to time-series classification datasets abundant with contextual information.  (2) GPT-2 is regarded as the base language model in our experimental evaluation. Incorporating recently prevalent  large language model~\cite{touvron2023llama,jin2023large} is very attractive compared to use these small-scale language model. (3) Although its effective, the proposed InstructTime largely ignore the collaboration between traditional deep learning models~\cite{zerveas2021transformer} and the language model. Harness the complementary strength of traditional time series classificier might be a potential extendation direction for improving our InstructTime.  


\clearpage  
\bibliographystyle{ACM-Reference-Format}
\bibliography{instructtime-sample-base}

\clearpage  
 \appendix
\section{Datasets Descriptions}
\label{sec:data}
In the next, we introduce the seven publicly available datasets, provided by~\cite{andrzejak2001indications,anguita2013public,cheng2023timemae,liu2021gated}. Particularly, we have meticulously curated text descriptions for the labels, drawing from authoritative sources and formal documentation associated with each dataset.
\vspace{-0.1in}
\begin{itemize}
	\item Sleep-EDF (Electroencephalogram, EEG): This dataset comprises 197 full-night Polysomnography (PSG) sleep recordings, sourced from PhysioNet. It facilitates the analysis of EEG signals, with labels categorizing sleep stages into Wake (W), Rapid Eye Movement (REM, denoted as R), stages one through four, Movement (M), and unscored segments, indicating unidentified stages.
	\item  PhysioNet Challenge 2020 (Electrocardiogram, ECG): This composite dataset is pivotal for the classification of cardiac abnormalities and includes sub-datasets from multiple cohorts (CPSC, CPSC-Extra, G12EC). The labels, delineating various cardiac conditions, are detailed on the challenge's official webpage.
	\item Human Activity Recognition (HAR): Sourced from 30 participants, this dataset captures six distinct activities—walking, ascending stairs, descending stairs, sitting, standing, and lying down—using a smartphone carried by the subjects, facilitating the recognition of human activities.
	\item Fault Detection in Electromechanical Drives (FD): The FD dataset is derived from an electromechanical drive system, focusing on the monitoring and detection of rolling bearing conditions. Labels include 'not damaged', 'inner damaged', and 'outer damaged', enabling precise fault diagnosis.
	\item Right Whale Call Detection (RWC): This dataset consists of audio recordings aimed at identifying the presence of right whale calls within a sea of ambient ocean sounds. Labels are binary, distinguishing between 'right whale' and 'unknown creature', to aid in the acoustic monitoring of marine life.
	\item Epilepsy: This dataset consists of time series data derived from electroencephalogram (EEG) recordings, which capture the electrical activity of the brain. This dataset is commonly used to develop and test algorithms for seizure detection and prediction, offering a valuable resource for understanding the patterns and anomalies associated with epileptic episodes.
	\item ArabicDigits (AD): It comprises audio recordings of spoken Arabic digits from zero to nine, captured from multiple native speakers. The dataset is structured as time series data representing the acoustic features of speech, making it an excellent resource for developing and evaluating models in speech recognition, particularly those aimed at understanding and classifying spoken numerical digits in Arabic.
\end{itemize}
Table~\ref{tab:datasets} in the supplementary material provides an extensive overview of each dataset, including specific details such as the division of data into training and test sets. For datasets inherently segmented into training and testing partitions (HAR, FD, RWC), we maintain the original distribution. Conversely, for datasets without predefined splits (Sleep-EDF, PhysioNet Challenge 2020), we allocate 90\% of the data for training and 10\% for testing~\cite{he2023shapewordnet}.

\begin{table}[ht]
	\centering
	\caption{Statistic of five used datasets in our experiments.}
	\vspace{-0.1in}
	\resizebox{0.475\textwidth}{!}{%
		\begin{tabular}{c|ccccc}
			\toprule
			Datasets & \# Train Size & \# Test Size & \# Length & \# Channel & \# Label \\
			\midrule
			EEG   & 12,787 & 1,421 & 3,000 & 2     & 8 \\
			ECG   & 10,854 & 1,206 & 5,000 & 12    & 27 \\
			HAR   & 8,823 & 2,947 & 128   & 9     & 6 \\
			FD    & 10,912 & 2,728 & 5,120 & 1     & 3 \\
			RWC   & 10,934 & 1,962 & 4,000 & 1     & 2 \\
			\hline
			ESR   & 9,200 & 2,300 & 178   & 1     & 2 \\
			AD    & 5,940 & 1,980 & 93    & 13    & 10 \\
			\bottomrule
		\end{tabular}%
	}
	\label{tab:datasets}%
\end{table}%

\section{Hyper-parameter Settings}
\label{sec:implement_appendix}

\subsection{Compared Baselines} 
In the appendix, we elaborate on the baseline models utilized in the main manuscript. These models span several methodologies, including self-attention-based architectures, convolutional networks, self-supervised learning frameworks, and an innovative adaptation of the GPT2 model for classification tasks. Below, we provide concise descriptions of each model, highlighting their principal techniques and conceptual foundations:

\begin{enumerate}
	\item \textbf{Self-attention based }:
	\begin{itemize}
		\item \textbf{TST}: Adapts the Transformer architecture for time series data, employing self-attention mechanisms to capture long-range dependencies within sequences.
		\item \textbf{Patch-TST}: Extends TST by segmenting time series into patches prior to applying the Transformer, enhancing efficiency in processing lengthy sequences by diminishing computational complexity.
		\item \textbf{FormerTime}: This model introduces modifications to the Transformer design, tailored for time series, integrating hierarchical multi-scale enhancements to boost performance.
	\end{itemize}
	
	\item \textbf{Convolutional based }:
	\begin{itemize}
		\item \textbf{MCDCNN}: This approach uses deep convolutional layers to extract multi-level features from multivariate time series, offering robust feature representation.
		\item \textbf{TCN }: TCNs leverage dilated convolutions to model long-term dependencies efficiently, providing a compelling alternative to recurrent architectures with reduced parameterization.
		\item \textbf{MiniROCKET}: MiniROCKET transforms time series into a discriminative feature set using convolutional kernels, enabling accurate classification.
	\end{itemize}
	
	\item \textbf{Self-supervised based }:
	\begin{itemize}
		\item \textbf{TimeMAE}: This model applies masked auto-encoding to time series, learning generic representations optimized by reconstructing original data from partially masked inputs with two loss terms.
		\item \textbf{TS-TCC}: TS-TCC employs contrastive learning to understand temporal dynamics, enhancing feature discriminability in an unsupervised manner.
	\end{itemize}
	
	\item \textbf{GPT-based}:
	\begin{itemize}
		\item \textbf{GPT-As-Classifier}:  This novel method modifies GPT2's final layer, leveraging its representative capacity for time series classification, making it adept at handling sequential data for time series classification.
	\end{itemize}
\end{enumerate}
Note that in our experiments, both Patch-TST and GPT-As-Classifier utilize identical input formats to ensure a fair comparison. For the self-supervised learning models, their architectures are based on a Transformer encoder network, with adjustments made to remove some factors that might lead to unfair advantages.

\subsection{Hyper-parameter Settings} We implement all models with Python 3.11 and PyTorch 2.1.2 in Nvidia GeForce RTX 4090. We leverage Adam as the optimizer with weight decay of $1e^{-5}$.  For all the baselines except those related to pre-trained language model (PLM), we consistently configure the model with an embedding size of $64$, a learning rate of $0.001$, and a batch size of $64$. 

Because the PLM has been pre-trained and the model itself has a large number of parameters compared to the baseline, our hyper-parameter settings are changed. For PLM, we always configured the model with the default number of layers and embedding sizes, a pre-training learning rate of $5e^{-5}$, a downstream task learning rate of $1e^{-5}$, and a batch size of $16$, and used the warm up with cosine annealing scheduler from the Transformers library, with a warm up ratio setting of $0.05$.

\section{Related Work}
\label{sec:related_work}
In this work, we mainly introduce two aspects of closely related work: time series classifier and pre-trained language model.
\subsection{Time Series Classification}
Time series classification (TSC) has garnered significant attention from researchers in recent years~\cite{middlehurst2023bake,shifaz2020ts}. Through an extensive review of the literature, these contributions can be broadly categorized into four types of classical approaches.  \textbf{Distance-based} methods form the bedrock of traditional TSC, primarily leveraging similarity measures to classify time series. The quintessential example is the dynamic time warping (DTW) algorithm, which, when combined with a nearest neighbor classifier (NN-DTW)~\cite{ding2008querying}, has been a benchmark in the field~\cite{petitjean2014dynamic}. \textbf{Interval-based} approaches focus on extracting features from specific intervals of the time series. Time series forest (TSF)~\cite{deng2013time} is a notable method in this category, employing random interval selection and summary statistics to capture local patterns. \textbf{ Shapelet-based}~\cite{ye2009time} methods revolve around identifying sub-sequences (shapelets) that are predictive of a time series class. The learning shapelets~\cite{grabocka2014learning} algorithm is a prime example, providing an automated process for shapelet discovery directly from the data.  \textbf{Dictionary-based} techniques involve transforming time series into symbolic representations and then analyzing the frequency of these symbolic patterns. One of the seminal works in this area is the Symbolic aggregate approXimation (SAX)~\cite{lin2007experiencing}, which enables a bag-of-words type model for TSC. This approach has been further extended by methods such as Bag of Patterns (BOP)~\cite{lin2012rotation} and symbolic aggregate approXimation - vector space model (SAX-VSM)~\cite{senin2013sax}, which have demonstrated efficacy in capturing repetitive and local patterns. Although effective in some scenarios, these methods excel in handling temporal distortions but can be computationally intensive for large datasets. In contrast,  \textbf{Deep Learning-based} methods have surged in popularity due to their ability to efficiently learn complex features from raw time series data~\cite{zhang2020tapnet,liu2024adaptive}.  Convolutional neural networks (CNNs)~\cite{zheng2014time,ismail2020inceptiontime}, and Self-attention based Transformer architectures~\cite{liu2021gated,cheng2023formertime} have been at the forefront of this wave, showcasing remarkable performance across diverse TSC tasks.  These deep learning methods benefit from the ability to model long-term dependencies and hierarchical feature representations without the need for manual feature engineering~\cite{christ2018time}. Despite the profound nonlinear modeling capabilities of deep neural networks and their advantage of obviating the need for manual feature engineering—thereby facilitating the learning of more complex temporal features for effective classification—their major drawback lies in their voracious data appetite. These models necessitate extensive labeled training sets, without which they are prone to overfitting. 
\subsection{Pre-trained Language Model}
In the realm of natural language processing (NLP), the advent of Pre-trained Language Models (PLM) has marked a paradigm shift from traditional static word vector representations, such as Word2Vec~\cite{mikolov2013efficient}, to dynamic, context-aware embeddings. PLM leverages vast amounts of textual data to learn rich, nuanced language representations before being fine-tuned for specific tasks, embodying a more holistic approach to language understanding. The evolution of PLM can be broadly categorized into three architectural paradigms: Decoder-only, Encoder-only, and Encoder-Decoder models. Each of these paradigms brings its unique approach to language modeling and representation. (1) Decoder-only models, such as GPT~\cite{radford2018improving}, focus on generating text based on the preceding context. This design is particularly adept at tasks involving language generation, where the model predicts subsequent tokens given a sequence of prior tokens. GPT and its successors exemplify this approach, demonstrating remarkable proficiency in text completion, creative writing, and more. (2) Encoder-only models, epitomized by BERT~\cite{devlin2018bert}, specialize in understanding and interpreting the context of a given text fragment. By processing text in a bidirectional manner, these models excel at tasks requiring deep contextual understanding, such as sentiment analysis, named entity recognition, and question answering. (3) Encoder-Decoder models, like BART~\cite{lewis2019bart}, combine the strengths of both encoders and decoders to handle a wide range of tasks from translation to summarization. These models are designed to encode the input text into an intermediate representation, which the decoder then uses to generate the output text, making them versatile tools for both understanding and generation tasks.

Subsequent research has explored the application of these PLM to downstream tasks, leveraging strategies such as Fine-tuning~\cite{devlin2018bert}, where the pre-trained model is slightly adjusted to perform specific tasks; Prompt engineering~\cite{brown2020language}, which frames tasks in a way that the model can understand; Instruction Tuning~\cite{ouyang2022training}, which adapts models to follow human-like instructions; and Reinforcement Learning from Human Feedback (RLHF), which refines models based on qualitative feedback. These methodologies have significantly expanded the utility of PLM, enabling their effective adaptation and application across a diverse spectrum of NLP challenges, thereby driving forward the frontiers of language understanding and generation.
\section{More Experimental Results} 
\label{sec:more_results}
\subsection{Study of Time Series Discretion}
\begin{table*}[htbp]
	\centering
	\caption{The influence of varying patch sizes during the tokenization stage in the InstructTime-Universal method.}
	\vspace{-0.1in}
	\resizebox{0.975\textwidth}{!}{%
		\begin{tabular}{c|ccccccccccccccc}
			\toprule
			\multirow{2}[2]{*}{Patch Size} & \multicolumn{3}{c}{EEG} & \multicolumn{3}{c}{ECG} & \multicolumn{3}{c}{HAR} & \multicolumn{3}{c}{FD} & \multicolumn{3}{c}{RWC} \\
			\cmidrule{2-16}                            & MSE   & Accuracy & F1 Score & MSE   & Accuracy & F1 Score & MSE   & Accuracy & F1 Score & MSE   & Accuracy & F1 Score & MSE   & Accuracy & F1 Score \\
			\midrule
			25,20,1,32,25 & \textbf{0.0534 } & 0.7889  & 0.4976  & \textbf{0.0227 } & 0.3735  & 0.2354  & \textbf{0.0501 } & \textbf{0.8695 } & \textbf{0.8683 } & \textbf{0.0597 } & 0.8702  & 0.8724  & \textbf{0.1941 } & 0.7113  & 0.7079  \\
			40,25,2,40,32 & 0.0595  & 0.7906  & 0.5085  & 0.0265  & 0.3569  & 0.2067  & 0.0557  & 0.6860  & 0.5933  & 0.0711  & \textbf{0.9205 } & \textbf{0.9314 } & 0.2024  & \textbf{0.7314 } & \textbf{0.7311 } \\
			50,30,4,64,40 & 0.0651  & \textbf{0.7931 } & \textbf{0.5122 } & 0.0296  & \textbf{0.3975 } & \textbf{0.2421 } & 0.0646  & 0.7486  & 0.6656  & 0.0851  & 0.8308  & 0.8516  & 0.2293  & 0.7021  & 0.7012  \\
			\bottomrule
		\end{tabular}%
	}
	\label{tab:patch_size}%
\end{table*}%
\begin{figure*}
	\centering
	\includegraphics[width=1.0\textwidth]{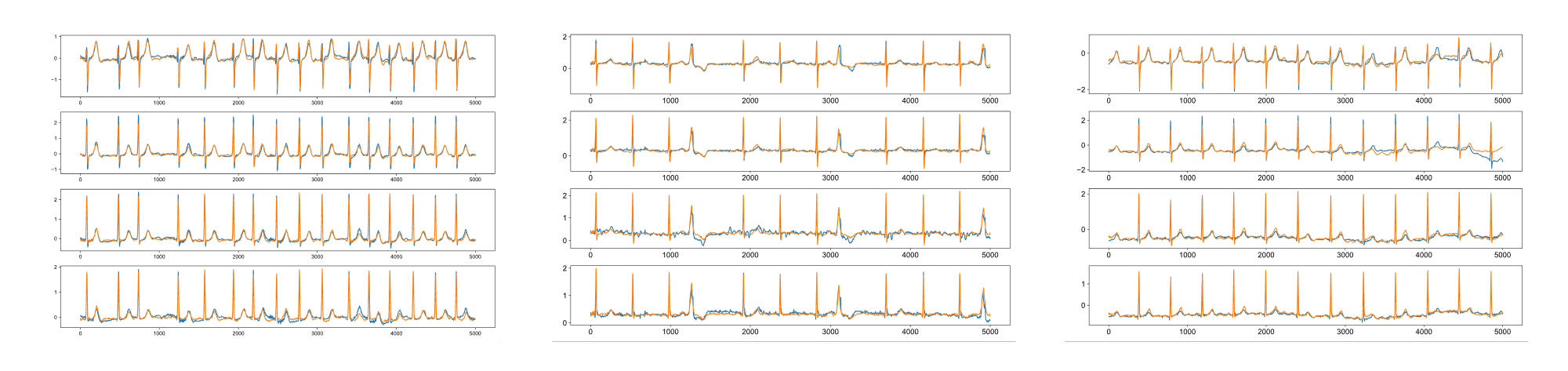}
	\vspace{-0.3in}
	\caption{Visualizing the reconstruction of ECG signals in the vector quantized networks.}
	\label{fig:reconstruction}
\end{figure*}
\subsubsection{Study of Varying Patch Size}
In this part, we introduce an experimental analysis to assess the impact of varying patch sizes on the performance of the InstructTime-Universal method across multiple datasets. Table \ref{tab:patch_size} presents the results of this investigation. Furthermore, the reconstructive results of VQ networks measured by the mean squared error (MSE) is also reported.  From the reported results, a evident trend is varying patch size have a non-uniform impact across different datasets. Notably, smaller patch sizes seem to yield better reconstruction performance while larger patch size correlate with improved classification performance for the FD and RWC datasets. This implies that while smaller patches can capture fine-grained details essential for signal reconstruction, larger patches may better encapsulate features relevant for classification tasks. It is also worth noting that there is no single patch size that outperforms others across all metrics and datasets, highlighting the need for tailored configurations to suit specific types of time series data. In summary, the results from Table \ref{tab:patch_size} indicate that patch size is a critical hyper-parameter in the InstructTime method that requires careful tuning to balance the trade-offs between reconstruction fidelity and classification performance. 
\subsubsection{Reconstructive Results of VQ Networks.}
Figure~\ref{fig:reconstruction} offers a comparative illustration of the original and reconstructed ECG signals within the Vector Quantized (VQ) networks. Each subplot presents two overlaid ECG traces: the original signal in blue and its reconstructed counterpart in orange. A visual inspection indicates that the VQ networks capture the fundamental morphology of the ECG time series effectively, with the P, QRS, and T waves being discernible in the reconstructed signal. However, there are noticeable deviations in certain segments where the reconstruction does not perfectly align with the original trace, particularly evident in the amplitude and sharpness of the R wave peaks. These discrepancies might be attributed to the quantization error inherent to the VQ networks, which can lead to a loss of fine detail in the signal. Despite these imperfections, the overall time series patterns and intervals  appear to be preserved. The consistency of the reconstructions across multiple cases suggests that the VQ networks have learned a generalized representation of the ECG data, which could be valuable to use in the multimodal modeling problem.

\subsection{Study of Instruction Data Construction}
\subsubsection{Importance of Sequence Length Consistency Across Domains}
In this part, we aim to examine the impact of sequence length consistency of multimodal prompt across various domains. Correspondingly, Table~\ref{tab:sequence_length} records the performance metrics for two sequence length variations across all datasets. From the results, we could find that our InstructTime model trained with a consistent sequence length across different datasets show a marked improvement.  For instance, the accuracy for the EEG dataset increased from $0.5852$ to $0.8067$, and similarly, the F1 score improved significantly.  This finding emphasizes the need for careful pre-processing of instruction data to ensure consistency in sequence length, which can be a critical factor in the successful deployment of models across various domains.
\begin{table*}[htbp]
	\centering
	\caption{Illustrating the importance of consistent sequence length of instruction data across different domains, evaluated in InstructTime-Universal.}
	\vspace{-0.1in}
	\resizebox{1\textwidth}{!}{%
		\begin{tabular}{c|cccccccccc}
			\toprule
			\multirow{2}[2]{*}{Sequence Length} & \multicolumn{2}{c}{EEG} & \multicolumn{2}{c}{ECG} & \multicolumn{2}{c}{HAR} & \multicolumn{2}{c}{FD} & \multicolumn{2}{c}{RWC} \\
			\cmidrule{2-11}                            & Accuracy & F1 Score & Accuracy & F1 Score & Accuracy & F1 Score & Accuracy & F1 Score & Accuracy & F1 Score \\
			\midrule
			200, 120, 128, 128, 125 & 0.5852 & 0.2754 & 0.2595 & 0.3438 & 0.8546 & 0.7900 & 0.8867 & 0.8895 & 0.6901 & 0.6592 \\
			125, 120, 128, 128, 125 & \textbf{0.8067} & \textbf{0.5007} & \textbf{0.3402} & \textbf{0.4820} & \textbf{0.8990} & \textbf{0.8944} & \textbf{0.9619} & \textbf{0.9656} & \textbf{0.7307} & \textbf{0.7299} \\
			\bottomrule
		\end{tabular}%
	}
	\label{tab:sequence_length}%
\end{table*}%

\subsubsection{Importance of Incorporating Label Texts into Prompt}
Here, we explore the influence of integrating textual descriptions of potential classification labels into the model prompts, as detailed in Table \ref{tab:textual_prompt}. Specifically, we compare the performance of model variants with and without the inclusion of label descriptions across four distinct datasets. From the reported results, we could find that the inclusion of label descriptions in prompts contributes positively to the model's performance. For instance, in the EEG dataset, the model's Accuracy increased from 0.7875 to 0.7931, and the F1 Score rose from 0.4879 to 0.5122 when label descriptions were included. This pattern of improved performance is consistent across all datasets. The consistent improvement across various datasets reaffirms the value of incorporating contextual information into the model prompts.
\begin{table*}[htbp]
	\centering
	\caption{Experimental results of incorporating textual descriptions of candidate classification labels into prompts, evaluated in InstructTime-Universal.}
	\vspace{-0.1in}
	\resizebox{0.975\textwidth}{!}{%
		\begin{tabular}{c|cccccccc}
			\toprule
			\multirow{2}[2]{*}{Model Variants} & \multicolumn{2}{c}{EEG} & \multicolumn{2}{c}{HAR} & \multicolumn{2}{c}{FD} & \multicolumn{2}{c}{RWC} \\
			\cmidrule{2-9}                            & Accuracy & F1 Score & Accuracy & F1 Score & Accuracy & F1 Score & Accuracy & F1 Score \\
			\midrule
			w/o Label Descriptions & 0.7875  & 0.4879  & 0.8405  & 0.8379  & 0.8712  & 0.8318  & 0.6934  & 0.6883  \\
			w/ Label Descriptions & \textbf{0.7931 } & \textbf{0.5122 } & \textbf{0.8695 } & \textbf{0.8683 } & \textbf{0.9245 } & \textbf{0.9301 } & \textbf{0.7314 } & \textbf{0.7311 } \\
			\bottomrule
		\end{tabular}%
	}
	\label{tab:textual_prompt}%
\end{table*}%
\begin{table*}[htbp]
	\centering
	\caption{Experimental results of different fine-tuning strategies in the stage of auto-regressive training, evaluated in InstructTime-Universal.}
	\vspace{-0.1in}
	\resizebox{0.975\textwidth}{!}{%
		\begin{tabular}{cc|cccccccccc}
			\toprule
			\multirow{2}[2]{*}{Strategies} & \multirow{2}[2]{*}{Model Variants} & \multicolumn{2}{c}{EEG} & \multicolumn{2}{c}{ECG} & \multicolumn{2}{c}{HAR} & \multicolumn{2}{c}{FD} & \multicolumn{2}{c}{RWC} \\
			\cmidrule{3-11}                       &       & Accuracy & F1 Score & Accuracy & F1 Score & Accuracy & F1 Score & Accuracy & F1 Score & Accuracy & F1 Score \\
			\midrule
			\multirow{2}[2]{*}{\begin{tabular}{@{}c@{}}w/o\\Cross-Domain\end{tabular}} & LoRA Fine-tuning & 0.7322 & 0.4210 & 0.1522 & 0.1685 & 0.6843 & 0.6625 & 0.8015 & 0.7148 & 0.6705 & 0.6589 \\
			& Full Finetuning & \textbf{0.7854} & \textbf{0.4854} & \textbf{0.2554} & \textbf{0.3751} & \textbf{0.8341} & \textbf{0.8296} & \textbf{0.9092} & \textbf{0.9203} & \textbf{0.7270} & \textbf{0.7268} \\
			\midrule
			\multirow{2}[2]{*}{\begin{tabular}{@{}c@{}}w/\\Cross-Domain\end{tabular}} & LoRA Fine-tuning & 0.7372 & 0.4002 & 0.2446 & 0.3326 & 0.8468 & 0.8435 & 0.8704 & 0.7306 & 0.6927 & 0.5665 \\
			& Full Finetuning & \textbf{0.8067} & \textbf{0.5007} & \textbf{0.3402} & \textbf{0.4820} & \textbf{0.8990} & \textbf{0.8944} & \textbf{0.9619} & \textbf{0.9656} & \textbf{0.7307} & \textbf{0.7299} \\
			\bottomrule
		\end{tabular}%
	}
	\label{tab:lora}%
\end{table*}%
\subsection{Study of Fine-tuning Strategies}
In this part, the main objective is to evaluate the efficacy of different fine-tuning strategies during the auto-regressive training of our model. Table~\ref{tab:lora} compares two fine-tuning approaches - LoRA fine-tuning and full fine-tuning - under two conditions: with and without access cross domain. From the results, we could clearly find that full fine-tuning consistently outperforms LoRA Fine-tuning across all datasets and in both with and without cross-domain scenarios. Meanwhile, the impact of cross-domain data is also notable, particularly in the full fine-tuning strategy, where the presence of cross-domain information leads to better generalization, as demonstrated by the higher performance metrics. Interestingly, while the LoRA Fine-tuning strategy does not match the performance of full fine-tuning, it still benefits from the inclusion of cross-domain data, suggesting that even more parameter-efficient fine-tuning methods can leverage additional data to improve model performance. To sum up, the reported results strongly advocate for full fine-tuning as the superior strategy in the auto-regressive training of models, particularly when combined with cross-domain data.

\subsubsection{\textbf{Convergence Curves Analysis}}
Figure 5 depicts two key aspects of the experimental analysis: the auto-regressive pre-training learning curves and the performance of various models on a downstream task with parameters frozen post-pre-training. The left part demonstrates a rapid decline in training loss within the initial epochs, indicating that the model quickly assimilates the fundamental structure of the training data. The subsequent epochs show a plateau in loss, suggesting that the model has reached a point of diminishing returns with regard to learning from the training dataset. On the right, the accuracy curves for various models on a downstream task are plotted. The ECG model maintains a consistently high performance across epochs, while the HAR model experiences a notable dip before stabilizing and improving. The FD, RWC, and ECO models start with lower accuracy and exhibit minimal improvement over time. This trend may point to the robustness of the ECG model in capturing the nuances of the downstream task, and possibly to the limitations of the other models when adapting to tasks outside their initial training scope. To sum up, the pre-training phase results in a model well-equipped for the downstream task, as evidenced by a rapid reduction in training loss and stable accuracy in frozen evaluation. The ECG model, in particular, demonstrates a strong transfer learning capability, maintaining high accuracy throughout the epochs, which underscores the potential utility of the pre-trained model in applications where model adaptation is constrained by frozen parameters.
\begin{figure}[t]
	\centering
	\includegraphics[width=0.5\textwidth]{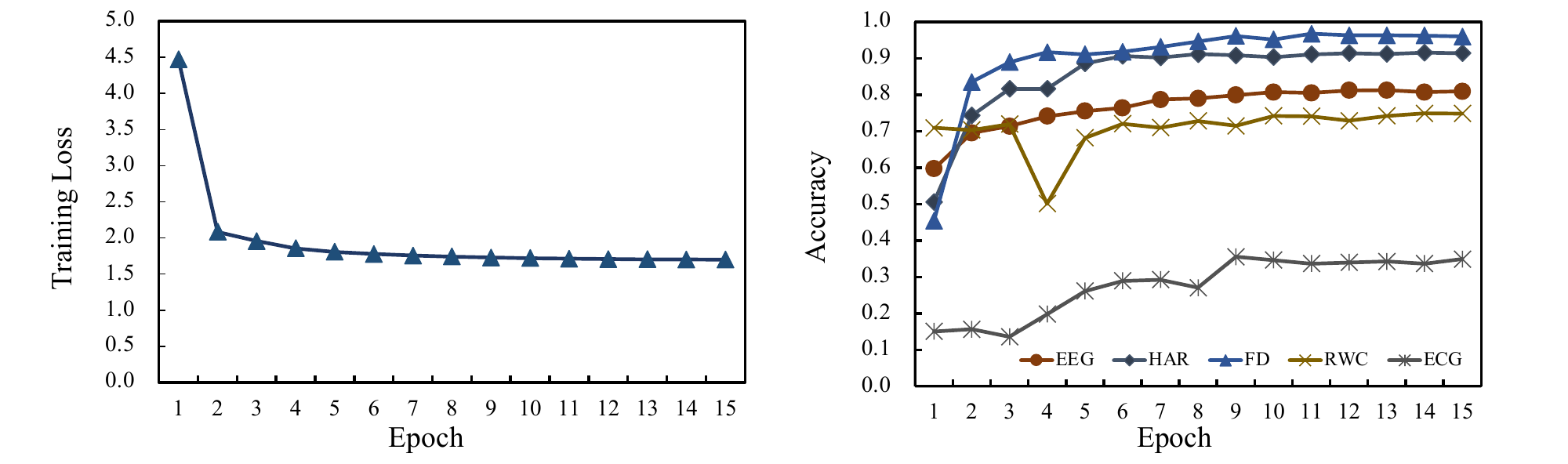}
	\vspace{-0.3in}
	\caption{Learning curves of auto-regressive pre-training and corresponding results of downstream task in terms of the InstructTime-Frozen.}
	\label{fig:learning_curves}
\end{figure}


\end{document}